\documentclass{article}

\usepackage{arxiv}

\usepackage[utf8]{inputenc} 
\usepackage[T1]{fontenc}    
\usepackage{hyperref}       
\usepackage{url}            
\usepackage{booktabs}       
\usepackage{amsfonts}       
\usepackage{nicefrac}       
\usepackage{microtype}      
\usepackage{lipsum}		
\usepackage{graphicx}
\usepackage{natbib}
\usepackage{doi}
\usepackage{amsmath}
\usepackage{tabularx}
\usepackage{multirow}

\title{The bionic neural network for external simulation of human locomotor system }

\author{ \href{https://orcid.org/0000-0000-0000-0000}{\includegraphics[scale=0.06]{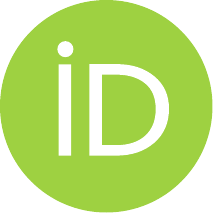}\hspace{1mm}Yue~Shi}\thanks{Yue Shi is with the School of Electronic and Electrical Engineering, University of Leeds, Leeds, UK. LS2 9JT (email:
y.shi1@leeds.ac.uk} \\
	Department of Electronic and Electrical Engineering \\
	University of Leeds\\
	Woodhouse Lane Leeds LS2 9JT \\
	\texttt{y.shi1@leeds.ac.uk} \\
	\And
	\href{https://orcid.org/0000-0000-0000-0000}{\includegraphics[scale=0.06]{orcid.pdf}\hspace{1mm} Shuhao~Ma} \\
	Department of Electronic and Electrical Engineering \\
	University of Leeds\\
	Woodhouse Lane Leeds LS2 9JT \\
	\texttt{1sma@leeds.ac.uk} \\
    \And
	\href{https://orcid.org/0000-0000-0000-0000}{\includegraphics[scale=0.06]{orcid.pdf}\hspace{1mm} Yihui~Zhao} \\
	Bristol robotic lab, \\
	University of Bristol,\\
	Bristol, UK. \\
	\texttt{yihui.zhao@bristol.ac.uk} \\
}

\renewcommand{\shorttitle}{\textit{arXiv} Template}

\hypersetup{
pdftitle={A template for the arxiv style},
pdfsubject={q-bio.NC, q-bio.QM},
pdfauthor={Yue Shi, Shuhao Ma, Yihui Zhao},
pdfkeywords={Electromyography (EMG), exoskeleton, online learning, adversarial learning, edge-computing, parallel computing},
}

\begin{document}
\maketitle

\begin{abstract}
Muscle forces and joint kinematics estimated with musculoskeletal (MSK) modeling techniques offer useful metrics describing movement quality. Model-based computational MSK models can interpret the dynamic interaction between the neural drive to muscles, muscle dynamics, body and joint kinematics, and kinetics. Still, such a set of solutions suffers from high computational time and muscle recruitment problems, especially in complex modeling. In recent years, data-driven methods have emerged as a promising alternative due to the benefits of flexibility and adaptability. However, a large amount of labeled training data is not easy to be acquired. This paper proposes a physics-informed deep learning method based on MSK modeling to predict joint motion and muscle forces. The MSK model is embedded into the neural network as an ordinary differential equation (ODE) loss function with physiological parameters of muscle activation dynamics and muscle contraction dynamics to be identified. These parameters are automatically estimated during the training process which guides the prediction of muscle forces combined with the MSK forward dynamics model. Experimental validations on two groups of data, including one benchmark dataset and one self-collected dataset from six healthy subjects, are performed. The results demonstrate that the proposed deep learning method can effectively identify subject-specific MSK physiological parameters and the trained physics-informed forward-dynamics surrogate yields accurate motion and muscle forces predictions. 

\end{abstract}

\keywords{Electromyography (EMG) \and exoskeleton \and online learning \and adversarial learning \and edge-computing \and parallel computing}

\section{Introduction}
Accurate estimation of muscle forces holds pivotal significance in diverse domains. As a powerful computational simulation tool, the musculoskeletal (MSK) model can be applied for detailed biomechanical analysis to understand muscle-generated forces thoroughly, which would be beneficial to various applications ranging from designing efficacious rehabilitation protocols \cite{1605264}, optimizing motion control algorithms \cite{5723418, RN95}, enhancing clinical decision-making \cite{8371283, KARTHICK201845, 8637973} and the performance of athletes \cite{9546647, app11041450}. Thus far, the majority of the MSK models are based on physics-based modeling techniques to interpret transformation among neural excitation, muscle dynamics, joint kinematics, and kinetics.\par

It is challenging to offer a biologically consistent rationale for the choice of any objective function \cite{https://doi.org/10.1002/jor.20876, MODENESE20112185}, given our lack of knowledge about the method used by the central nervous system (CNS) \cite{RN96}. Moreover, estimation of muscle forces and joint motion during inverse dynamics analysis suffers from high computational time and muscle recruitment problems, especially in complex modeling \cite{TRINLER201955, 4912337, doi:10.1080/23335432.2014.993706}. The surface electromyography (sEMG) signal is a non-invasive technique that measures the electrical activity of muscles. The sEMG signal is able to reflect the motor intention of the human 20 ms to 200 ms prior to the actual initiation of the joint motion and muscle activation \cite{10.3389/fneur.2020.00934, 8004462}. MSK forward dynamics model is an effective method for mapping sEMG to muscle force and joint motion. The sEMG-based forward dynamics approaches calculate the muscle forces based on the sEMG signals and muscle properties which can directly reflect the force and velocity of muscle contraction without assuming optimization criteria or constraints. For instance, Zhao et al. proposed an MSK model driven by sEMG for estimating the continuous motion of the wrist joint and a genetic algorithm for parameters optimization \cite{9258965}. Thomas et al. presented an MSK model to predict joint moment and related muscle forces of knee and elbow\cite{Buchanan2004NeuromusculoskeletalME}. Although MSK forward dynamics approaches are effective, access to individualised physiological parameters is challenging. Optimisation algorithms are used commonly for parameter identification which needs the muscle forces or joint moments calculated from inverse dynamics as the targets. However, the process of obtaining the targets is time-consuming \cite{RN84, 10.1371/journal.pone.0216663}. To address the time-consuming and uncertainty issues of model-based methods, data-driven methods have also been explored to establish relationships between movement variables and neuromuscular status, i.e., from sEMG to joint kinematics and muscle forces \cite{9380441, 10.1371/journal.pone.0247883}. Burton et al. \cite{BURTON2021110439} implemented four machine/deep learning methods, including recurrent neural network (RNN), convolutional neural network (CNN), fully-connected neural network (FNN), and principal component regression, to predict trend and magnitude of the estimated joint contact and muscle forces. Zhang et al. proposed a physics-informed deep learning framework for the prediction of the muscle forces and joint Kinematics where a new loss function from motion equation was designed as soft constraints to regularize the data-driven model \cite{9970372}. Shi et al. seamlessly integrated Lagrange's equation of motion and inverse dynamic muscle model into the generative adversarial network (GAN) framework for sEMG-based estimation of muscle force and joint kinematics \cite{shi2023physicsinformed}. 

However, despite their adaptability, and flexibility, deep learning methods still suffer from potential limitations. It is a ’black box’ method, the intermediate processes of data-driven models do not consider the physical significance underlying the modeling process. More importantly, the above deep learning methods used labeled input and output data for training and it is difficult for us to get the value of the muscle forces corresponding to the sEMG. Given the considerations of the limitations concerning both model-based and data-driven models, in this work, we propose a novel PINN framework combining the information from MSK forward dynamics for the estimation of joint angle and related muscle forces, simultaneously identifying the physiological parameters. We find that the entire MSK forward dynamics system can be regarded as an ordinary differential equation (ODE) system with unknown physiological parameters, where the independent variable is the joint angle. Specifically, a fully connected neural network (FNN) estimates the MSK forward dynamics model. The physiological parameters are identified during back-propagation. In addition, muscle forces are predicted through the guide of muscle contraction dynamics.

The main contributions of this paper include: The PINN framework is proposed for the estimation of joint motion and related muscle forces, simultaneously identifying the physiological parameters. In the absence of true values, muscle forces are estimated based on embedded muscle contraction dynamics through sEMG signals and joint motion signals.


\section{METHODS}\label{method}
In this section, an introduction to each sub-process of the MSK forward dynamics system is first presented in Part. A. Then, we demonstrate a novel PINN framework in Part. B. 

\subsection{Musculoskeletal forward dynamics}\label{1}
\subsubsection{Muscle activation dynamics}\label{Muscle activation dynamic}
Muscle activation dynamics refer to the process of transforming sEMG signals $e_n$ into muscle activation signals $a_n$ which can be estimated through Eq. \ref{eq}.
\begin{equation}
a_n = \frac{e^{Ae_n} - 1}{e^A - 1}
\label{eq}
\end{equation}
where preprocessed sEMG signals $e_n$ are utilized as neural activation signals $u_n$ for estimating muscle activation signals \cite{6226835} to simplify the model. $A$ is a nonlinear factor which has the range of highly non-linearity (-3) to linearity (0.01) \cite{RN78}.

\subsubsection{Muscle-Tendon Model}\label{2}
\begin{figure}
  \centering
  \includegraphics[scale = 0.4]{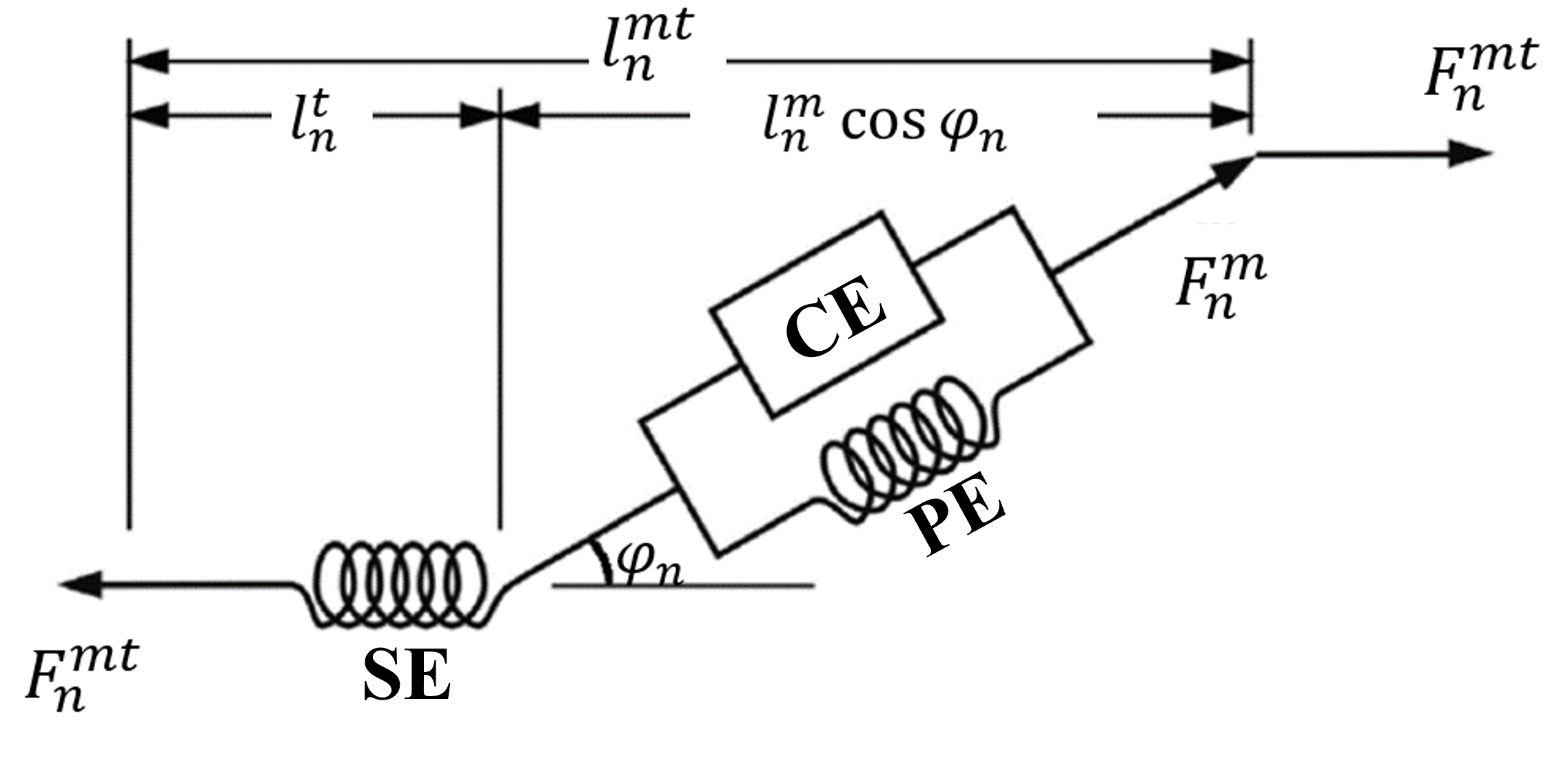} 
  \caption{Wrist model with the primary muscle units associated with wrist flexion and extension}
  \label{fig:0}
\end{figure}

Hill’s modelling technique is used to compute the muscle-tendon force $F^{mt}$, which consists of an elastic tendon in series with a muscle fibre. The muscle fibre includes a contractile element (CE) in parallel with a passive elastic element (PE), as shown in Fig. \ref{fig:0}. $l^{mt}_n, l^{m}_n, l^{t}_n$ is the muscle-tendon length, muscle fibre length and tendon length of the nth muscle respectively. 
The model is parameterized by the physiological parameters maximum isometric muscle force $F_{0, n}^{m}$, the optimal muscle fibre length $l_{0, n}^{m}$, the maximum contraction velocity $v_{0, n}$, the slack length of the tendon  $l_{s, n}^{t}$ and the pennation angle  $\varphi_{0, n}$, which are difficult to measure in vivo and varies between the age, gender. We employ the following vector $ \chi_n $ to represent the personalized physiological parameters of the nth muscle that require identification.
$$ \chi_n = [l^m_{0,n}, v_{0,n}, F_{0,n}^{m}, l_{s,n}^{t}, \varphi_{0,n}] $$

Pennation angle $\varphi_n$ is the angle between the orientation of the muscle fibre and tendon, and the pennation angle at the current muscle fibre length $l^{m}_n$ is calculated by
\begin{equation}
\varphi_n(l^{m}_n; \chi_n) = \sin^{-1}(\frac{l^m_{0,n}\sin{\varphi_{0,n}}}{l^{m}_n})
\label{fi}
\end{equation}
In this study, the tendon is assumed to be rigid \cite{10.1115/1.4023390} and thus the tendon length $l_{s,n}^{t} = l^{t}_n$ is adopted. The consequent fibre length can be calculated as follows:
\begin{equation}
l^m_{n} = (l^{mt}_n - l^{t}_n){\cos^{-1}{\varphi_n}}
\label{lm}
\end{equation}

The $F_{CE, n}$, is the active force generated by CE, which can be written as
\begin{equation}
F_{CE,n}(a_n, l^{m}_n, v_n;\chi_n) = a_nf_{v}(\overline{v}_{n})f_{a}(\overline{l}_{n,a}^{m})F_{0,n}^{m}
\label{Fce} 
\end{equation}
The function $f_a(\cdot)$ represents the active force-length relationship at different muscle fibre lengths and muscle activations, which are written as
\begin{equation}
f_{a}(\overline{l}_{n,a}^{m}) = e^{{-({\overline{l}}_{n,a}^{m}-1)}^{2}k^{-1}}
\label{Fce}
\end{equation}
where the ${\overline{l}}_{n, a}^{m} = {l}^{m}_{n} / ({l}_{0, n}^{m}(\lambda(1 - a_n) + 1)$ is the normalized muscle fibre length concerning the corresponding activation levels and $\lambda$ is a constant, which is set to 0.15 \cite{LLOYD2003765}. The $k$ is a constant to approximate the force-length relationship, which is set to 0.45 \cite{10.1115/1.1531112}. The function $f_{v}(\overline{v}_{n})$ represents the force-velocity relationship between the $F^m_n$ and the normalized contraction velocity $\overline{v}_{n}$ \cite{article}
\begin{equation}
\label{Fv}
\begin{aligned}
f_{v}(\overline{v}_{n}) &= \left\{\begin{array}{ll}
\frac{0.3(\overline{v}_n+1)}{-\overline{v}_n+0.3} & \overline{v}_n \leq 0 \\
\frac{2.34 \overline{v}_n + 0.039}{1.3\overline{v}_n + 0.039} & \overline{v}_n > 0\\
\end{array}\right.
\end{aligned}
\end{equation}
where $\overline{v}_{n} = v_n / v_{0,n}$. $v_{0, n}$ is typically set as 10 ${l}_{0, n}^{m}$/sec \cite{Zajac1989MuscleAT}. The $v_n$ is the derivative of the muscle fibre length with respect to time $t$. 

Note that the passive force $F_{PE,n}$ is the force produced by PE which can be calculated as
\begin{equation}
\label{Fv}
\begin{aligned}
 F_{PE,n}(l^{m}_n;\chi_n)&= \left\{\begin{array}{ll}
\ 0 & {l}_{n}^{m} \leq {l}_{0,n}^{m} \\
\ f_{P}(\overline{l}^{m}_{n})F_{0,n}^{m} & {l}_{n}^{m} > {l}_{0,n}^{m}\\
\end{array}\right.
\end{aligned}
\end{equation}
where $\overline{l}^{m}_{n} = {l}_{n}^{m} / {l}_{0,n}^{m}$ the normalized muscle fibre length. The $f_p(\cdot)$ is
\begin{equation}
f_{P}(\overline{l}^{m}_{n}) = \frac{e^{10(\overline{l}^{m}_{n} - 1)}}{e^5}
\label{fp}
\end{equation}

The $F^{mt}_n$ is the summation of the active force $F_{CE,n}$ and the
passive force $F_{PE,n}$, which can be written as
\begin{equation}
\begin{aligned}
F^{mt}_n(a_n, l^{m}_n, v_n; \chi_n) = (F_{CE,n} + F_{PE,n})\cos{\varphi_n}\\
\label{Fmt}
\end{aligned}
\end{equation}

\subsubsection{Joint Kinematic Modelling Technique}\label{3}
The single joint configuration is presented to estimate the wrist's continuous joint motion. The muscle-tendon length $l^{mt}_n$ and moment arm $r_n$ against wrist joint angle $q$ are calculated using the polynomial equation and the scale coefficient \cite{RAMSAY2009463}. The total joint torque is calculated as 
\begin{equation}
\tau(\mathbf{a}, \mathbf{r}, \mathbf{l}^{m}, \mathbf{v}; \tilde{\chi}) =\sum_{n=1}^{N} F_n^{mt}(a_n, l^{m}_n, v_n;\chi_n)r_n
\label{torque}
\end{equation}
Where N represents the number of muscles included. $\mathbf{a}, \mathbf{r}, \mathbf{l}^{m}, \mathbf{v}, \tilde{\chi}$ represent the sets of muscle activation, muscle arm length, muscle fibre length, muscle contraction velocity and the  physiological parameters of all included muscles, respectively.
Since the muscle activation level is not directly related to the joint motion, it is necessary to compute joint acceleration using forward dynamics. In this research, we only consider the simple single hinge joint's flexion and extension and take the model of the wrist joint's flexion as an example.
The wrist joint is assumed to be a single hinge joint, and the palm and fingers are assumed to be a rigid segment rotating around the wrist joint in the sagittal plane. Thus, we can have the following relationship based on the Lagrange equation
\begin{equation}
\ddot{q} = \frac{\tau - mgL\sin{q} - c\dot{q}}{I}
\label{Lagrange equation}
\end{equation}
where $q, \dot{q}, \ddot{q}$ are the joint motion, joint angular velocity and joint angular acceleration. $I$ is the moment of inertia of hand, which is equal to $mL^{2} + I_{p}$. $I_{p}$ is the moment of inertia at the principal axis which is parallel to the flexion/extension axis. $m$ and $L$ are the mass of the hand and the length between the rotation centre to the hand’s centre of mass, which are measured from subjects. $C$ is the damping coefficient representing the elastic and viscous effects from tendons and ligaments. $\tau$
calculated from Eq. \ref{torque}.
The three parts mentioned above constitute the main components of the MSK  forward dynamics model, which establishes the process from sEMG to joint angle motion from a model-based method. 

The reason why the model-based approach is introduced here is because it is the foundation on which the PINN framework is built later. We embedded the entire MSK forward dynamics approach into the neural network

The reason why the model-based method is introduced here is because this is the basis on which the PINN framework is built. We embed the entire MSK forward dynamics method into the neural network
\subsection{PINN framework}
\begin{figure*}
  \centering
  \includegraphics[width=0.8\textwidth]{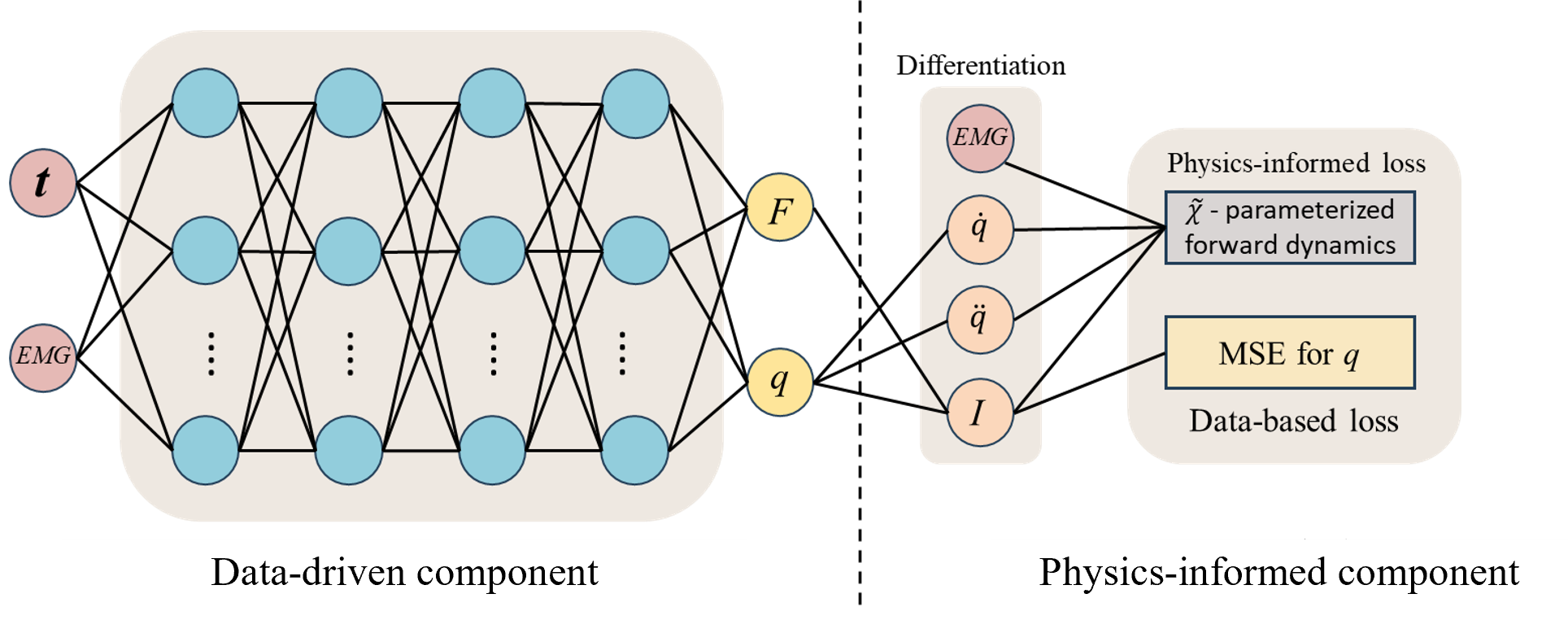} 
  \caption{A computational graph of the proposed PINN.}
  \label{fig:1}
\end{figure*}

The proposed PINN framework for the estimation of joint angle and muscle forces together with physiological parameters identification is introduced. The computational graph is illustrated in Fig. \ref{fig:1}.
To be specific, in the data-driven component, a Fully connected Neural Network (FNN) is utilised to automatically extract the high-level features and build the relationship between sEMG signals and the joint angle/muscle forces, while the physics-informed component entails the underlying physical relationship between the joint angle and muscle forces.
In this manner, in the data-driven component, the recorded sEMG signals and discrete time steps $t$ are first fed into FNN. With the features extracted by FNN and integration of the MSK forward dynamics model, the predicted muscle forces and joint angles could be achieved.
\subsubsection{Architecture and Training of FNN}
To demonstrate the effectiveness of the proposed physics-informed deep learning framework, we chose an architecture of FNN which was composed of four fully-connected blocks and two regression blocks. Specifically, each the fully-connected block has one Linear layer, one ReLU layer and one dropout layer. Similar to the fully-connected blocks, there are one ReLU layer and one dropout layer in each regression block. Between these two regression blocks, one is for the prediction of joint angle, the other is for the prediction of muscle forces. The training was performed using the Adam algorithm with an initial learning rate of $1 * 10^{-4}$. 
In the model training phase, the batch size is set as 1, and FNN is trained by stochastic gradient descent with momentum. Additionally, the maximum iteration is 1000 and the dropout rate is 0.3.

\subsubsection{Design of Loss Functions}
Unlike state-of-the-art methods, the loss function of the proposed framework consists of the MSE loss and the physics-based losses. The MSE loss is 
to minimise the mean square error of the ground truth and prediction, while the physics-informed losses are for the parameters identification and guidance on the muscle force prediction. The total loss shows below:
\begin{equation}
L_{total} = L_q + L_{r,1} + L_{r,2}
\end{equation}
\begin{equation}
L_q = MSE(q)
\end{equation}
\begin{equation}
L_{r,1} = R_1(q)
\end{equation}
\begin{equation}
L_{r,2} = R_2(q)
\end{equation}

where $L_q $ denotes the MSE loss function of the joint angle since it has the ground truth from measurement, while $L_{r,1}$ represents the first residual loss function imposed by the MSK forward dynamics introduced in Sec. \ref{1}, which helps identify physiological parameters simultaneously. $L_{r,2}$ denotes the second residual loss function which is utilized for the prediction of muscle forces. $R_1(q), R_2(q)$ both indicate the function of the predicted joint angle. The mathematical formulas for each part of the loss will be specified next.
\paragraph{MSE Loss} MSE loss is calculated by
\begin{equation}
MSE(q) = \frac{1}{T} \sum_{t=1}^{T}(\hat{q_t} - q_t)^2
\end{equation}
where $q_t$ is denoted as the ground truth of the joint angle at time step t, and $\hat{q_t}$ is the predicted joint angle from the network. Additionally, $T$ denotes the total sample number.

\paragraph{Physics-informed Loss} Physics-informed governing laws, reflecting underlying relationships among the muscle force and kinematics in human motion, are converted to constraints during the Neural Network's training phase. 
The first residual loss function designed from the Lagrange equation which expressed the motion state of the MSK system is shown below
\begin{equation}
\begin{aligned}
R_1(q) &= \frac{1}{T} \sum_{t=1}^{T} (I\ddot{\hat{q_t}} + C\dot{\hat{q_t}} + mgL\sin{\hat{q_t}} \\
&- \tau_t(\mathbf{a}, \mathbf{r}, \mathbf{l}^{m}, \mathbf{v}^{m}; \tilde{\chi}))  
\end{aligned}
\end{equation}

$\tau_t$ represents the joint torque which is calculated by Eq. \ref{torque}. The moment arm $\mathbf{r}$ against the joint angle is calculated using the polynomial equation and the scale coefficient \cite{RAMSAY2009463}, which can be represented as $\mathbf{r}(q)$. We use $\mathbf{l}^{mt}(q)$ to represent the relationship between joint angle and the muscle-tendon length which is exported from Opensim. Therefore, muscle fibre length $\mathbf{l}^m(q)$ can also be expressed in terms of the joint angle $q$ from Eq. \ref{lm}. The $\mathbf{v}(q)$ is the derivative of the muscle fibre length $\mathbf{l}^m(q)$, which can be represented in terms of $q, \ddot{q}$. 
As mentioned in Eq. \ref{Lagrange equation}, this equation expresses the dynamic equilibrium equation that the dynamical system should satisfy at time $t$. Therefore, for the MSK forward dynamics surrogate model from the neural network, this equation is still satisfied. When the predicted angle of the neural network is ${\hat{q_t}}$ at time $t$, the joint torque at this time can be expressed as $\tau_t(\mathbf{a}, \mathbf{r}(\hat{q_t}), \mathbf{l}^{mt}(\hat{q_t}), \mathbf{v}^{m}(\hat{q_t}); \tilde{\chi})$ = $\tau_t(\mathbf{a}, \hat{q_{t}}, \dot{\hat{q_{t}}}; \tilde{\chi})$, which can be regarded as an ordinary differential equation (ODE) with unknown physiological parameters $\tilde{\chi}$ in respect of the predicted joint angle $\hat{q_t}$. During the training phase, $\dot{\hat{q_t}}, \ddot{\hat{q_t}}$ can be automatically derived through backpropagation.
These physiological parameters $\tilde{\chi}$ are defined as the weights of the surrogate neural network which are then automatically updated as the neural network minimises the residual loss function $L_{r,1}$ during backpropagation. The estimation of $\hat{\chi}$ can be written as
\begin{equation}
\hat{\chi} = \underset{\tilde{\chi}}{\arg \min} L_{r,1}
\label{para}
\end{equation}

Due to the time-varying nature and difficulty in measurement of muscle forces, it is challenging for us to get the measured muscle forces as the ground truth for training. The second residual loss function $L_{r,2}$ is designed for the guidance of the muscle forces prediction which is based on $L_{r,1}$. During each iteration, we identify physiological parameters through $L_{r,1}$. These potential physiological parameters are used to calculate the muscle forces through the MSK forward dynamics model embedded as the learning target of the neural network. The loss function $L_{r,2}$ is given by
\begin{equation}
R_2(q) = \frac{1}{T} \sum_{t=1}^{T}\sum_{n=1}^{N}(\hat{F^{mt}_{t,n}} - F^{mt}_{t,n}(a_{t,n}, l^{m}_{t,n}, v^{m}_{t,n}; \hat\chi_{n}))
\end{equation}
where $\hat{F^{mt}_{t,n}}$ is the estimated muscle force from the network at the time step $t$. $F^{mt}_{t,n}$ represents the muscle force calculated by Eq. \ref{Fmt}. $l^{m}_{t,n}, v^{m}_{t,n}$indicates the muscle fibre length and muscle contraction speed of the nth muscle at time $t$. They all can be expressed in terms of the $q, \dot{q}$. $\hat\chi_{n}$ stands for the identified physiological parameters for the nth muscle by Eq. \ref{para}. Specifically, muscle force of the nth muscle at time $t$ can be expressed as follows $F^{mt}_{t,n}(a_{t,n}, \hat{q_{t}}, \dot{\hat{q_{t}}}; \hat\chi_{n})=F^{mt}_{t,n}(a_{t,n}, l^{m}_{t,n}, v^{m}_{t,n}; \hat\chi_{n})$ which can be regarded as another ODE with respect to the joint angle $\hat{q_{t}}$ as well, guiding the muscle forces prediction in the absence of the ground truth. 

\section{DATASET AND EXPERIMENTAL METHODS}\label{Dataset}
A self-collected dataset of wrist motions is utilised to demonstrate the feasibility of the proposed framework.
\subsubsection{Selection and initialisation of physiological parameters to be identified}
Among the physiological parameters of the muscle-tendon units involved, we only chose the maximum isometric muscle force $F_{0, n}^{m}$ and the optimal muscle fibre length $l_{0, n}^{m}$ for the identification in order to increase the generality of the model. The nonlinear shape factors $A$ in the activation dynamics also need to be identified.  Physiological parameters other than these were obtained by linear scaling based on the initial values of the generic model from Opensim. The selection and initialisation of all the physiological parameters are summarized table. \ref{tab:table0}. Since there may be differences in terms of magnitude and scale between each parameter because of their different physiological nature, it is necessary to deflate them in about the same interval before training the network.
\begin{table}[h]
\scriptsize
\centering
\caption{Physiological parameters involved in the forward dynamics setup of wrist flexion-extension motion for subject 1}
\resizebox{0.5\textwidth}{!}{
\begin{tabular}{@{}c*{6}{c}@{}}
\toprule[\heavyrulewidth]
Parameters       & FCR   & FCU   & ECRL  & ECRB  & ECU    \\ \midrule
$F_{0, n}^{m}$(N)   & 407   & 479   & 337   & 252   & 192    \\
$l_{0, n}^{m}$(m)   & 0.062 & 0.051 & 0.081 & 0.058 & 0.062  \\
$v_{0, n}$(m/s)       & 0.62  & 0.51  & 0.81  & 0.58  & 0.62   \\
$l_{s, n}^{t}$(m)   & 0.24  & 0.26  & 0.24  & 0.22  & 0.2285 \\
$\varphi_{0, n}$(rad) & 0.05  & 0.2   & 0     & 0.16  & 0.06   \\
\midrule
A                & \multicolumn{5}{c}{0.01}               \\ 
\bottomrule[\heavyrulewidth]
\end{tabular}
}
\label{tab:table0}
\end{table}

\subsubsection{Self-Collected Dataset}
Approved by the MaPS and Engineering Joint Faculty Research Ethics Committee of the University of Leeds (MEEC18-002), 6 subjects participated in this experiment. The consent forms are signed by all subjects. We take the subject’s weight data and the length of their hand in order to calculate the moment of inertia of the hand.

In the experiment, subjects were informed to maintain a fully straight torso with the $90$ abducted shoulder and the $90$ flexed elbow joint. The continuous wrist flexion/extension motion was recorded using the VICON motion capture system. The joint angle was computed at 250 Hz through the upper limb model using 16 reflective markers. Meanwhile, sEMG signals were recorded by Avanti Sensors at 2000 Hz from the main wrist muscles, including FCR, FCU, ECRL, ECRB and ECU. Moreover, the sEMG signals and motion data were synchronised and re-sampled at 1000 Hz. 5 repetitive trials were performed for each subject, and a three-minute break was given between trials to prevent muscle fatigue.

The measured sEMG signals were band-pass filtered (20 Hz and 450 Hz), fully rectified, and low-pass filtered (6 Hz). Then, they were normalised concerning the maximum voluntary contraction recorded before the experiment, resulting in the enveloped sEMG signals. Each wrist motion trial, consisting of time steps $t$, pre-processed sEMG signals and wrist joint angles, is formed into at by 7 matrix.

\subsubsection{Evaluation Criteria}
To quantify the estimation performance of the proposed framework, root mean square error (RMSE) is first used as the metric. RMSE is the response variable. Specifically, RMSE indicates the discrepancies in the amplitude and between the estimated variables and the ground truth, which can be calculated by
\begin{equation}
RMSE = \sqrt{{\frac{1}{T}\sum_{t=1}^{T}(\hat{y_{t}} - y_{t}})^2}
\end{equation}
where $y_{t}$ and $\hat{y_{t}}$ indicate the ground truth and the corresponding predicted value, respectively.

The coefficient of determination which denoted $R^{2}$ is used as another metric, which could be calculated by 
\begin{equation}
\label{r2}
R^2 = 1 - \frac{\sum_{i=1}^{T}(y_t - \hat{y}_t)^2}{\sum_{t=1}^{T}(y_t - \bar{y_t})^2}
\end{equation}
Where $T$ is the number of samples, $y_t$ is the measured ground truth at the time $t$, $\hat{y}_t$ is the corresponding estimation of the model, and $\bar{y}$ is the mean value of all the samples.

\subsubsection{Baseline Methods and Parameters Setting}
To verify the effectiveness of the proposed physics-informed deep learning framework in the prediction of the muscle forces and joint angle, several state-of-the-art methods, including FNN, SVR \cite{8930581}, are considered as the baseline methods for the comparison. Specifically, FNN has the same neural network architecture as our proposed model which has four fully-connected blocks and two regression blocks but without the physics-informed component. Stochastic gradient descent with Adam optimiser is employed for FNN training, the batch size is set as 1, the maximum iteration is set as 1000, and the initial learning rate is 0.001. The radial basis function (RBF) is selected as the kernel function of the SVR method and the parameter $C$ which controls the tolerance of the training samples is set as 100 and the kernel function parameters $\gamma$ which controls the range of the kernel function influence is set to 1.  

\section{RESULTS}\label{results}

In this section, we evaluate the performance of the proposed framework on a self-collected dataset which is detailed in Sec. \ref{Dataset} by comparing it with selected baseline methods.
Specifically, we first present the process of parameter identification by our model and verify the reliability of the identification results. Then, the overall comparisons depict the outcomes of both the proposed framework and the baseline methods. This includes representative results showcasing the included muscles' force and joint angles, alongside comprehensive predictions for six healthy subjects.
In addition, we investigate the robustness and generalization performance of the proposed framework in the intrasession scenario. 
Finally, the effects of the network architectures and hyperparameters on our model are evaluated separately. 
The proposed framework and baseline methods are trained using PyTorch on a laptop with GeForce RTX 3070 Ti graphic cards and 32G RAM.
\subsection{Parameters Identification}

\begin{figure*}[t]
    \centering
    \includegraphics[width=0.8\textwidth]{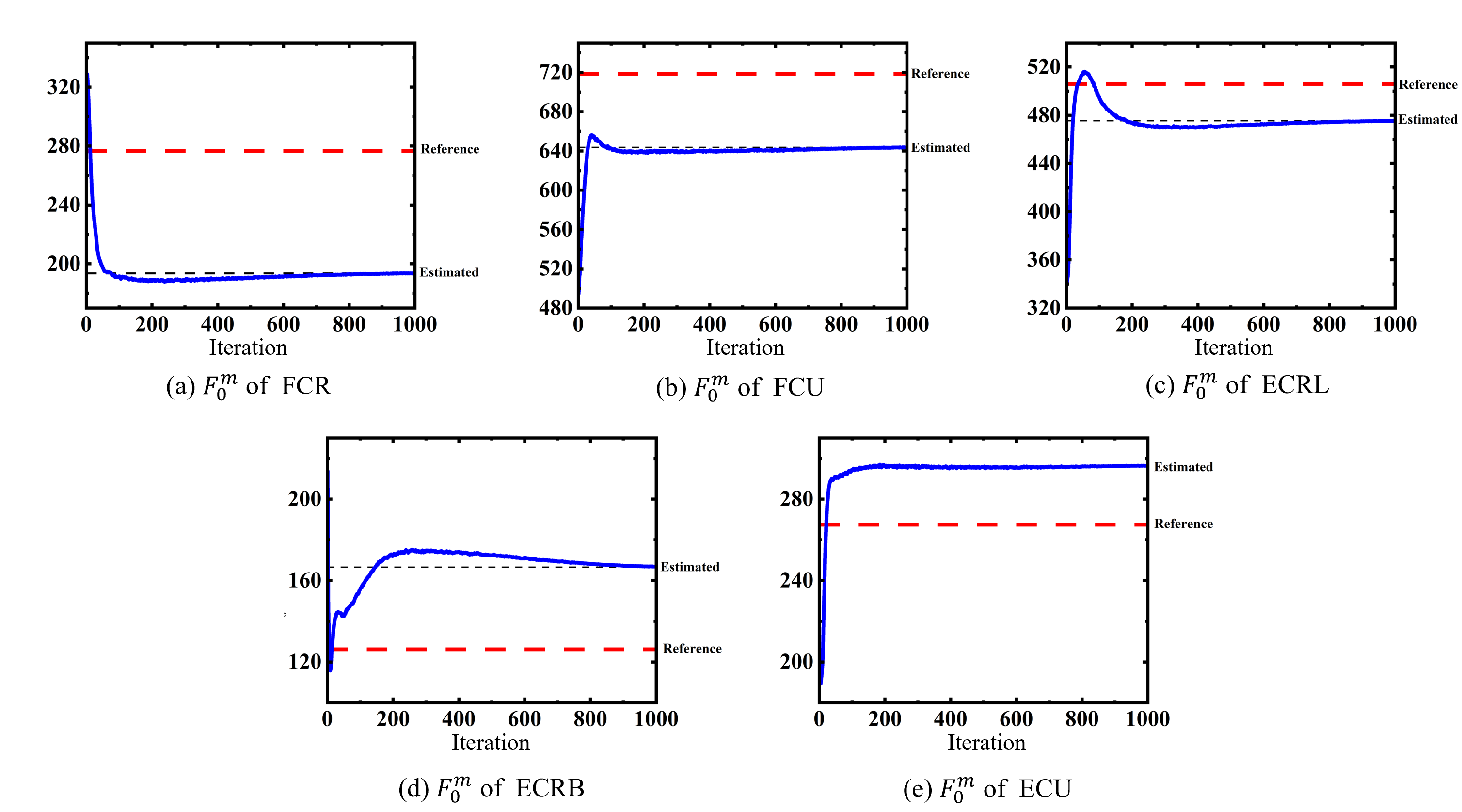}
    \caption{Evolution of the maximum isometric muscle force $F_0^m$ identified during the training process of subject 1.}
    \label{fig:2}
\end{figure*}

\begin{figure*}[t]
    \centering
    \includegraphics[width=0.8\textwidth]{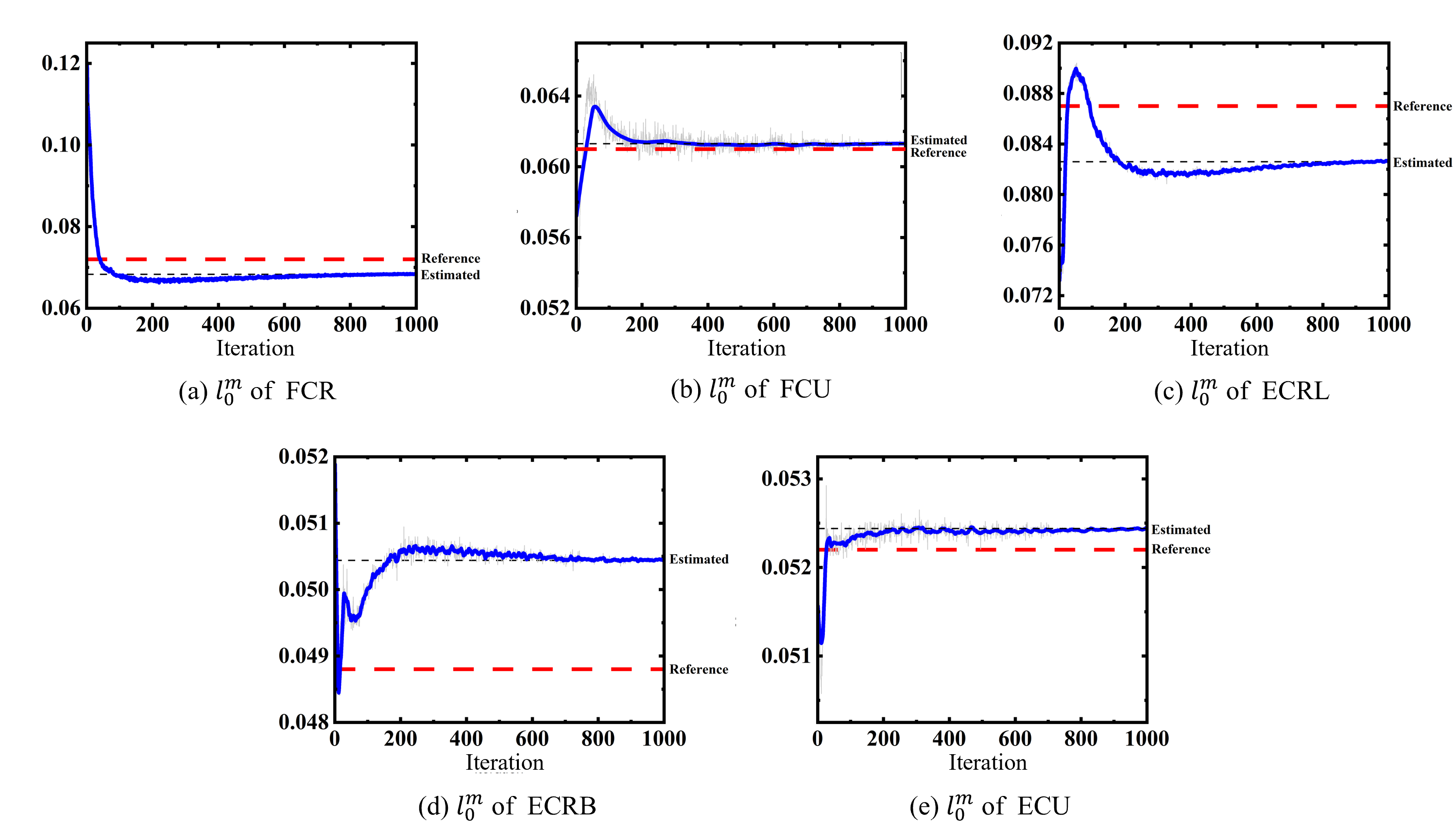}
    \caption{Evolution of the optimal muscle fiber length $l_0^m$ identified during the training process of subject 1.}
    \label{fig:2}
\end{figure*}

The subject-specific physiological parameters are identified during the training. Table. \ref{tab:table1} presents the estimation and the variation of the parameters of subject one. The variation pertains to the extent of disparity between the initial guess and the determined value derived from our model.
The initial guess of the parameters is given in Table. \ref{tab:table0} in Sec. \ref{method}. Physiological boundaries of the parameters are chosen according to \cite{RN86}. The boundaries of maximum isometric force $F^{m}_{0}$ are set to $50\%$ of the initial guess, while the boundaries of optimal muscle fibre length $l^{m}_{0}$ are set to $\pm 0.001$ of the initial guess. As shown in the Table. \ref{tab:table1}, the parameters identified through our proposed framework are all within the physiological range and possess physiological consistency. Furthermore, the deviations of the optimal fibre length exhibit minimal disparity from the initial value, while the maximum isometric forces diverge significantly from the initial value.
As well as the two main muscle-tendon physiological parameters, the identified non-linear shape factor A muscle activation dynamics parameter A is -2.29 which is physiologically acceptable in the range of -3 to 0.01.
\begin{table}[htbp]
\centering
\caption{Identified physiological parameters of Subject one}
\begin{tabularx}{0.5\textwidth}{l *{4}{X}}
\toprule
             & \multicolumn{4}{c}{Parameter index}                                                                                 \\ \cmidrule(lr){2-5}
             & \multicolumn{2}{c}{${l}_{0}^{m}$(m)}                                    & \multicolumn{2}{c}{${F}_{0}^{m}$(N)}                                    \\ \cmidrule(lr){2-3} \cmidrule(lr){4-5}
Muscle index & Estimation                   & Variation                    & Estimation                   & Variation                    \\ 
\midrule
FCR          & 0.056                     & 90.78\%                      & 475.2                     & 116.71\%                     \\
FCU          & 0.061                    & 120.62\%                     & 644.1                     & 129.37\%                     \\
ECRB         & 0.050                    & 86.15\%                      & 166.7                     & 66.05\%                      \\
ECRL         & \multicolumn{1}{l}{0.082} & \multicolumn{1}{l}{101.85\%} & \multicolumn{1}{l}{475.2} & \multicolumn{1}{l}{140.91\%} \\
ECU          & \multicolumn{1}{l}{0.052} & \multicolumn{1}{l}{84.51\%}  & \multicolumn{1}{l}{286.6} & \multicolumn{1}{l}{148.58\%} \\ 
\bottomrule
\end{tabularx}
\label{tab:table1}
\end{table}
Fig. \ref{fig:2} shows the evolution of the physiological parameters identified during the training process. In conjunction with the comparison of results obtained through the PINN framework within physiological ranges, we establish the credibility of the results from another perspective.
We set the estimation of the physiological parameters from the MSK forward dynamics model optimized by Genetic Algorithm (GA) as a reference for comparison \cite{9258965}. As depicted in Fig. \ref{fig:2}, the red dashed lines represent the reference values estimated by the GA optimization method, the black dashed lines indicate the estimates from our proposed framework, and the blue solid lines illustrate the variation process of the parameters. The comparison revealed small differences between the values gained from two different methods. 
\subsection{Overall Comparisons}
\begin{figure*}
  \centering
  \includegraphics[scale=1]{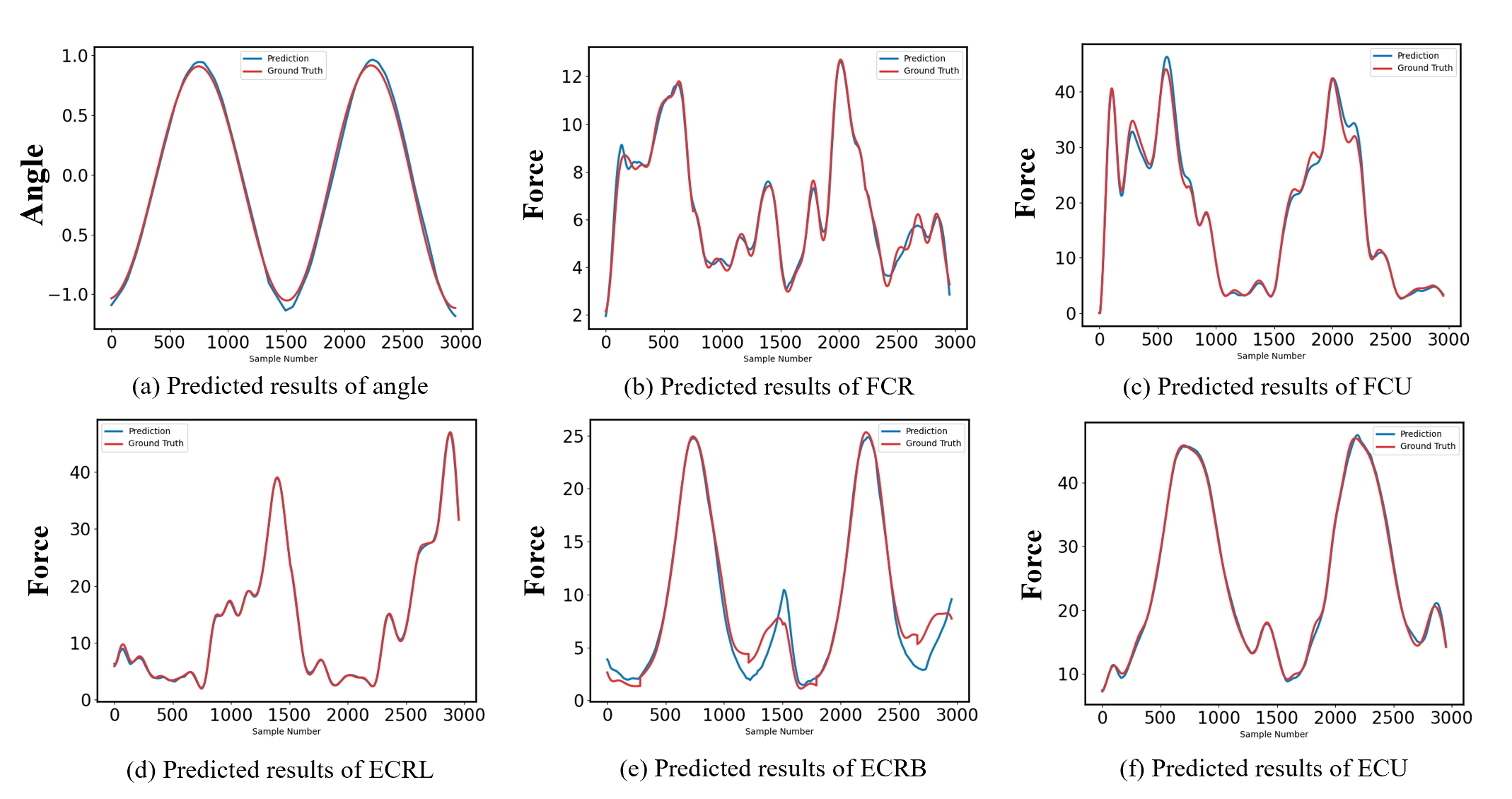} 
  \caption{Representative results of the wrist joint through the proposed data-driven model. The predicted outputs include the wrist angle, FCR muscle force, FCU muscle force, ECRL muscle force, ECRB muscle force, and ECU muscle force.}
  \label{fig:3}
\end{figure*}

\begin{table*}
\centering
\caption{RMSE AND R$^2$ of the proposed framework and baseline methods of wrist joint}
\resizebox{0.7\textwidth}{!}{
\begin{tabular}{cccccccc}
\toprule
Subject & Methods & FCR(N) & FCU(N) & ECRL(N) & ECRB(N) & ECU(N) & Angle(rad) \\
\midrule
\multirow{3}{*}{S1} & Ours & 4.99/0.98 & 4.62/0.97 & 4.35/0.98 & 5.33/0.96 & 3.12/0.96 & 0.09/0.98 \\
                    & FNN & 3.27/0.99 & 5.37/0.96 & 4.20/0.98 & 4.29/0.98 & 3.03/0.96 & 0.09/0.98 \\
                    & SVR & 6.02/0.97 & 5.59/0.96 & 6.05/0.96 & 4.50/0.98 & 5.21/0.94 & 0.11/0.97 \\
\midrule
\multirow{3}{*}{S2} & Ours & 6.01/0.96 & 4.47/0.97 & 3.71/0.98 & 4.95/0.97 & 2.56/0.97 & 0.08/0.98 \\
                    & FNN & 5.57/0.97 & 4.69/0.97 & 2.79/0.99 & 5.76/0.96 & 2.04/0.98 & 0.06/0.99 \\
                    & SVR & 7.35/0.95 & 5.58/0.96 & 5.93/0.96 & 6.01/0.96 & 4.31/0.96 & 0.09/0.98 \\
\midrule
\multirow{3}{*}{S3} & Ours & 5.43/0.98 & 3.74/0.98 & 4.91/0.97 & 3.91/0.98 & 3.20/0.96 & 0.13/0.96 \\
                    & FNN & 6.25/0.97 & 4.27/0.98 & 5.29/0.97 & 5.59/0.96 & 2.97/0.96 & 0.13/0.96 \\
                    & SVR & 7.63/0.95 & 6.91/0.96 & 7.97/0.95 & 6.25/0.96 & 3.21/0.94 & 0.14/0.96 \\
\midrule
\multirow{3}{*}{S4} & Ours & 5.54/0.97 & 5.21/0.97 & 5.31/0.97 & 3.58/0.97 & 3.92/0.97 & 0.04/0.99 \\
                    & FNN  & 5.38/0.97 & 5.82/0.96 & 5.57/0.97 & 4.77/0.96 & 4.28/0.96 & 0.08/0.98 \\
                    & SVR  & 6.27/0.96 & 6.35/0.96 & 6.45/0.96 & 5.42/0.96 & 5.39/0.95 & 0.10/0.97 \\
\midrule
\multirow{3}{*}{S5} & Ours & 5.41/0.96 & 5.23/0.97 & 3.98/0.98 & 3.87/0.97 & 4.41/0.95 & 0.11/0.97 \\
                    & FNN & 4.57/0.97 & 4.61/0.97 & 4.53/0.97 & 4.01/0.97 & 3.82/0.96 & 0.14/0.96 \\
                    & SVR & 8.65/0.95 & 5.34/0.97 & 6.16/0.96 & 5.79/0.96 & 5.75/0.94 & 0.18/0.94 \\
\midrule
\multirow{3}{*}{S6} & Ours & 3.21/0.99 & 3.27/0.98 & 5.61/0.97 & 6.28/0.94 & 3.97/0.95 & 0.06/0.99 \\
                    & FNN & 3.97/0.98 & 5.11/0.97 & 5.50/0.97 & 6.24/0.94 & 4.32/0.94 & 0.05/0.99 \\
                    & SVR & 7.09/0.94 & 6.19/0.96 & 7.15/0.95 & 5.72/0.95 & 4.97/0.93 & 0.07/0.98 \\
\bottomrule
\end{tabular}
}
\label{table:0}
\end{table*}

\begin{figure}
\centering
\includegraphics[scale=0.6]{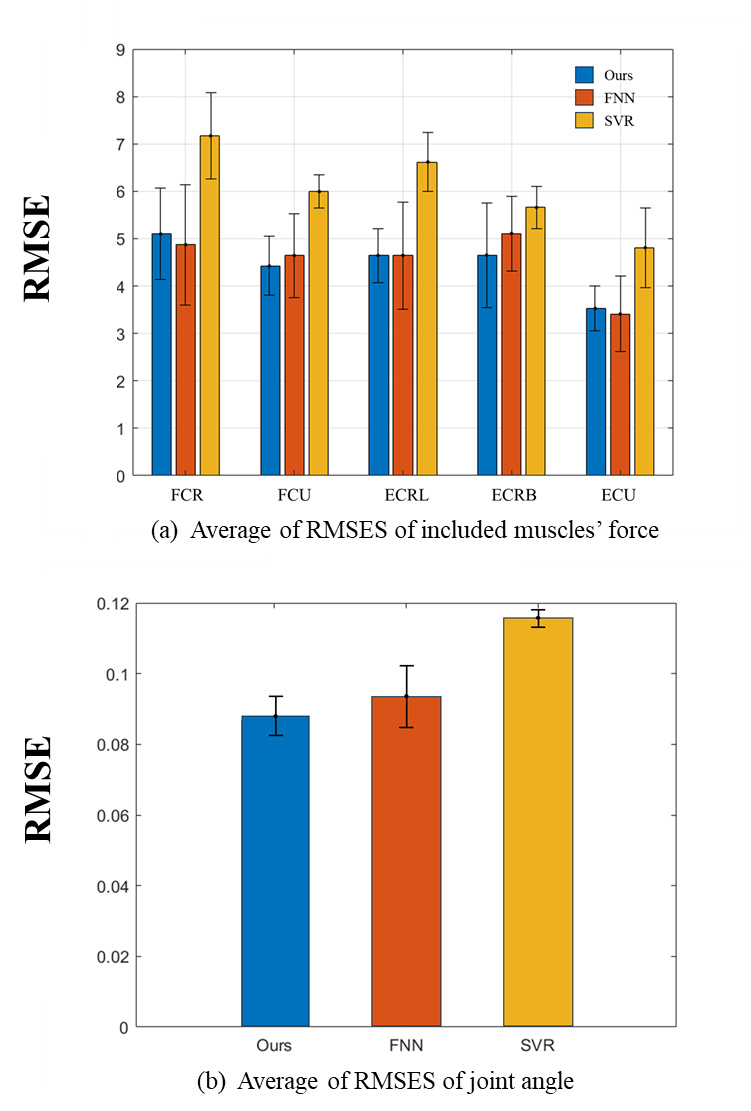}
\caption{Average RMSEs across all the subjects in (a) included muscles' force and (b) joint angle, respectively.}
\label{fig:5}
\end{figure}
The overall comparisons between the proposed framework and baseline methods are then performed. Fig. \ref{fig:3} depicts the representative results of the proposed framework for the wrist joint prediction, including the wrist angle, muscle force of FCR, muscle force of FCU, muscle force of ECRL, muscle force of ECRB, and muscle force of ECU, respectively. For the prediction of the joint angle, we use the measured value as the ground truth. 
However, for the prediction of muscle forces, since it is challenging to measure directly, we use the muscle forces calculated from the sEMG-driven MSK forward dynamics model as the ground truth\cite{article}.
As presented in Fig. \ref{fig:3}, the predicted values of muscle forces and joint angle can fit the ground truths well, indicating the great dynamic tracking capability of the proposed framework. 

To quantitatively evaluate the performance of our model, detailed comparisons of all the subjects between the PINN and the baseline methods are presented in Table. \ref{table:0}. We use the data with the same flexion speed to train and test the proposed framework and baseline methods in the wrist joint, according to Table. \ref{table:0}, deep learning-based methods, including the proposed framework, FNN, achieve better-predicted performance than the machine learning-based method, the SVR, as evidenced by smaller RMSEs and higher $R^2$ in most cases. Because these deep learning-based methods could automatically extract high-level features from the collected data. 

Fig. \ref{fig:5} (a) illustrates the average RMSEs of the muscle forces of the PINN framework and baseline methods across all the subjects. Among deep learning-based methods, our proposed framework achieves an overall performance similar to that of the FNN without direct reliance on actual muscle force labels. Fig. \ref{fig:5} (b) illustrates the average RMSEs of the joint angle. Our model achieves satisfactory performance with lower standard deviations, and its predicted results are with smaller fluctuations compared with the pure FNN method and the machine learning method. In the training process, the labelled joint angles are not only trained by the MSE loss but also enhanced by the physics-informed losses. The embedded physics laws provide the potential relationships between the output variables as learnable features for the training process. 

\subsection{Evaluation of Intrasession Scenario}
\begin{figure*}
  \centering
  \includegraphics[scale=1]{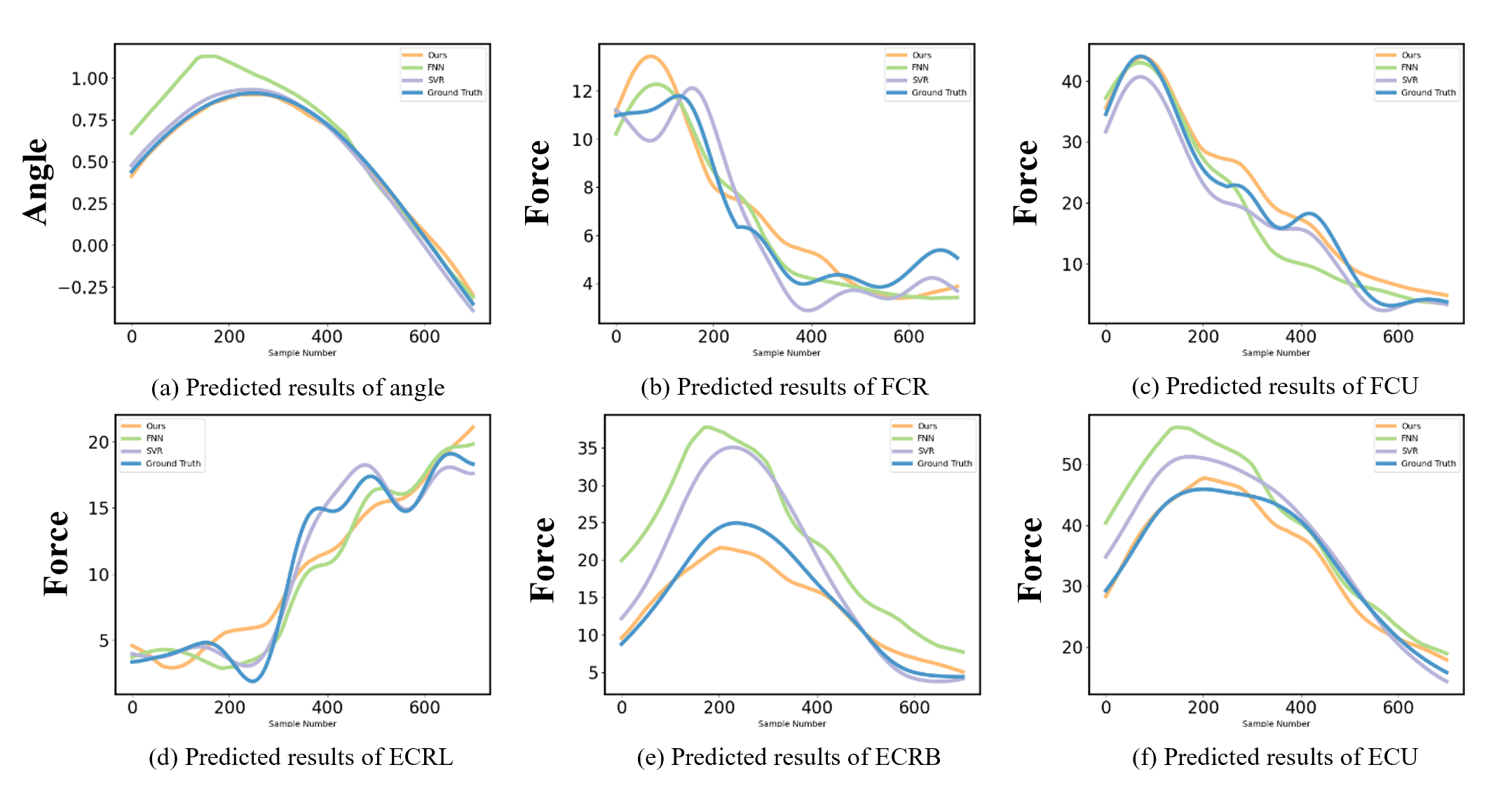} 
  \caption{ Evolution of the physiological parameters identified during the training process.}
  \label{fig:4}
\end{figure*}
The intrasession scenario is also considered to validate the robustness of the proposed framework. 
For each subject, we train the model with data of one flexion speed and then test the performance using data of a different flexion speed from the training data which can be seen as a dataset that our trained model is unfamiliar with. 
Fig. \ref{fig:4} depicts the corresponding experimental results, Our model demonstrates exceptional performance in datasets with different distributions, but the predicted results of some baseline methods are degraded. In particular, concerning the prediction results of the muscle forces of the ECRB and ECU, the prediction results yielded by our proposed framework demonstrate a notably enhanced congruence with the underlying ground truth. 
The SVR method also demonstrates its efficacy in accurately tracking specific predictions, such as the prediction of the muscle forces of the FCU and ECRL. Nevertheless, errors remain in other prediction results. Hence, the application of this method is constrained by various factors, including the nature of the prediction target, leading to both instability and inherent limitations.
The SVR method also exhibits the proficiency of dynamical tracking in part of the prediction results, such as the prediction of the muscle forces of the FCU and ECRL. However, the error remains in other prediction results which reflects the limitation of the stability. 
Although the FNN method shows favourable performance on the training data, when it comes to the intrasession scenario, the performance got much worse than that on the training data.
Our proposed framework manifests a discernible capability to predict muscle forces and joint angles on data characterized by the diverse distribution in the absence of the real value of the muscle forces.


\subsection{Effects of Hyperparameters}
To investigate the effects of hyperparameters, including learning rate, types of activation functions, and batch size, on the performance of the proposed framework, the detailed results are shown in Table. \ref{table:3}, Table. \ref{table:4}, and Table. \ref{table:5}. Specifically, we consider three different learning rates, i.e., 0.01, 0.001 and 0.0001, and the maximum iteration is set as 1000. According to Table. \ref{table:3}, we can find that the proposed framework achieves better performance with a smaller learning rate. As observed in Table. \ref{table:4}, the proposed framework with ReLU and Tanh could achieve comparable performance, but better than the one using Sigmoid as the activation function. In Table. \ref{table:5}, the performance of the proposed framework achieves better performance with the decrease of the batch size and achieves its best performance when the batch size is 1. Even though setting the batch size to 1 for training the deep learning framework can still retain the drawbacks such as over-fitting, poor generalisation, unstable training, etc., in previous experiments, they have been proven not to be issues.


\begin{table}[h]\scriptsize%
\centering
\caption{Comparisons of the proposed framework under various learning rates (R$^2$)}
\resizebox{0.5\textwidth}{!}{
\begin{tabular}{@{}c*{6}{c}@{}}
\toprule
Learning Rates & FCR & FCU & ECRL & ECRB & ECU & Angle \\ \midrule
0.01           & 0.96   & 0.93  & 0.95    & 0.94    & 0.93    & 0.96     \\
0.001          & 0.98   & 0.97  & 0.96    & 0.97    & 0.97    & 0.98    \\
0.0001         & 0.98   & 0.96  & 0.94    & 0.97    & 0.97    & 0.96     \\ 
\bottomrule
\end{tabular}
}
\label{table:3}
\end{table}

\begin{table}
\centering
\caption{Comparisons of the proposed framework under various Activation Functions (R$^2$)}
\resizebox{0.5\textwidth}{!}{
\begin{tabular}{@{}c*{6}{c}@{}}
\toprule
Activation Functions & FCR & FCU & ECRL & ECRB & ECU & Angle \\ \midrule
Sigmoid              & 0.16   & 0.21  & 0.15    & 0.17    & 0.09    & 0.20 \\
Tanh                 & 0.26   & 0.20  & 2.20    &  0.17   & 0.30    & 0.33     \\
ReLU                 & 0.98   & 0.97  & 0.96    & 0.97    & 0.97    & 0.98      \\
\bottomrule
\end{tabular}
}
\label{table:4}
\end{table}

\begin{table}[h]\scriptsize%
\centering
\caption{Comparisons of the proposed framework under various Batch Sizes (R$^2$)}
\resizebox{0.5\textwidth}{!}{
\begin{tabular}{@{}c*{6}{c}@{}}
\toprule
Batch Sizes & FCR & FCU & ECRL & ECRB & ECU & Angle \\ \midrule
1             & 0.98    & 0.97    & 0.96     & 0.97     & 0.97    & 0.98     \\
16            & 0.83    & 0.97    & 0.96     & 0.90     & 0.97    & 0.96     \\
32            & 0.82    & 0.96    & 0.96     & 0.91     & 0.97    & 0.90    \\ \bottomrule
\end{tabular}
}
\label{table:5}
\end{table}

\section{LIMITATIONS AND FUTURE DIRECTION}\label{discussions}
Although our proposed framework has shown promising performance in some scenarios, it still has shortcomings. In this section, we will discuss the current limitations and how future work can address them from five perspectives.

Our proposed framework has only been validated for single-joint, single-degree-of-freedom motion currently. In future research, the model's accuracy will be validated in complex motions involving multiple joints and degrees of freedom. 
More complex dynamical system modelling is also required for more complex motions. More parameters included in complex dynamic models are a big challenge for the identification through neural networks as well.

Our proposed framework primarily integrates physics laws from the MSK forward dynamics model as soft constraints for neural networks which are beneficial for the parameters identification and the guidance of the joint angle and muscle force prediction. Future work will incorporate additional physical constraints, such as environmental constraints, energy, physiological, etc. to construct more loss functions based on this physical information to enhance the performance of parameter identification and the physiological consistency of prediction results.

Our proposed model can perform well on preset fixed-period motions and it is difficult to generalise it for the effective prediction of other motions where the complexity and variability of the movement patterns are. Specifically, it can predict motions that approximate motion patterns, but for random and irregular motion, Our proposed model is almost completely incapable of making valid predictions. To solve this problem, in future research, Changing the form of our input data is a possible way. For example, abandon the use of the time-series sequences which consist of sEMG signals, time, and joint motion as the input samples. Alternatively, a cycle of motion can be used as a sample. Apart from altering the form of the input data, changing into a neural network that is more suitable for processing time series sequences, such as Recurrent Neural Network (RNN) and Long Short-Term Memory (LsTM) is another possible solution for the problem.

For our proposed framework, the number of physiological parameters that can be efficiently identified is limited
considering the network's performance and the time cost of model training. However, a more complex but physiologically consistent model with more parameters will directly benefit the accuracy of the muscle force prediction in our proposed model. Therefore, in future work, we will devoted to enhancing the model's performance to effectively identify a broader range of physiological parameters which will enable a deeper integration of the MSK forward dynamics model into our framework.

In order to improve the experimental feasibility and generalisation of the model, we partially simplified the MSK forward dynamics model by reducing the number of individualised physiological parameters and by choosing a rigid rather than an elastic tendon. A more physiologically accurate representation of muscle tissues with connective tissues and muscle fibres could inform the muscles' length and velocity-dependent force generation capacity. In addition, We will achieve precise physiological parameter identification and muscle force prediction on more accurate, complex MSK models in the future.

\section{CONCLUSION}\label{conclusion}
This paper presents a novel physics-informed deep-learning framework for joint angle and muscle force estimation. 
Specifically, our proposed model uses the ODE based on the MSK forward dynamics as the residual loss for the identification of personalized physiological parameters and another residual constraint based on the muscle contraction dynamics for the estimation of muscle forces in the absence of the ground truth. Comprehensive experiments on the wrist joint for muscle forces and joint angles' prediction indicate the feasibility of the proposed framework. 
However, it is worth noting that the entire MSK forward dynamics model embedded in the framework needs to be adjusted when the proposed approach is extended to other application cases.

\bibliographystyle{unsrtnat}
\bibliography{reference}  

\begin{thebibliography}{35}
\providecommand{\natexlab}[1]{#1}
\providecommand{\url}[1]{\texttt{#1}}
\expandafter\ifx\csname urlstyle\endcsname\relax
  \providecommand{\doi}[1]{doi: #1}\else
  \providecommand{\doi}{doi: \begingroup \urlstyle{rm}\Url}\fi

\bibitem[Dong et~al.(2006)Dong, Lu, Sun, and Rudolph]{1605264}
Shufang Dong, Ke-Qian Lu, Jian~Qiao Sun, and K.~Rudolph.
\newblock Adaptive force regulation of muscle strengthening rehabilitation device with magnetorheological fluids.
\newblock \emph{IEEE Transactions on Neural Systems and Rehabilitation Engineering}, 14\penalty0 (1):\penalty0 55--63, 2006.
\newblock \doi{10.1109/TNSRE.2005.863839}.

\bibitem[Ding et~al.(2010)Ding, Hirasawa, Kurita, Takemura, Takamatsu, Mizoguchi, and Ogasawara]{5723418}
Ming Ding, Kotaro Hirasawa, Yuichi Kurita, Hiroshi Takemura, Jun Takamatsu, Hiroshi Mizoguchi, and Tsukasa Ogasawara.
\newblock Pinpointed muscle force control in consideration of human motion and external force.
\newblock In \emph{2010 IEEE International Conference on Robotics and Biomimetics}, pages 739--744, 2010.
\newblock \doi{10.1109/ROBIO.2010.5723418}.

\bibitem[Luo et~al.(2023)Luo, Androwis, Adamovich, Nunez, Su, and Zhou]{RN95}
Shuzhen Luo, Ghaith Androwis, Sergei Adamovich, Erick Nunez, Hao Su, and Xianlian Zhou.
\newblock Robust walking control of a lower limb rehabilitation exoskeleton coupled with a musculoskeletal model via deep reinforcement learning.
\newblock \emph{Journal of NeuroEngineering and Rehabilitation}, 20\penalty0 (1):\penalty0 34, 2023.
\newblock ISSN 1743-0003.
\newblock \doi{10.1186/s12984-023-01147-2}.
\newblock URL \url{https://doi.org/10.1186/s12984-023-01147-2}.

\bibitem[Azimi et~al.(2020)Azimi, Nguyen, Sharifi, Fakoorian, and Simon]{8371283}
Vahid Azimi, Thang~Tien Nguyen, Mojtaba Sharifi, Seyed~Abolfazl Fakoorian, and Dan Simon.
\newblock Robust ground reaction force estimation and control of lower-limb prostheses: Theory and simulation.
\newblock \emph{IEEE Transactions on Systems, Man, and Cybernetics: Systems}, 50\penalty0 (8):\penalty0 3024--3035, 2020.
\newblock \doi{10.1109/TSMC.2018.2836913}.

\bibitem[Karthick et~al.(2018)Karthick, Ghosh, and Ramakrishnan]{KARTHICK201845}
P.A. Karthick, Diptasree~Maitra Ghosh, and S.~Ramakrishnan.
\newblock Surface electromyography based muscle fatigue detection using high-resolution time-frequency methods and machine learning algorithms.
\newblock \emph{Computer Methods and Programs in Biomedicine}, 154:\penalty0 45--56, 2018.
\newblock ISSN 0169-2607.
\newblock \doi{https://doi.org/10.1016/j.cmpb.2017.10.024}.
\newblock URL \url{https://www.sciencedirect.com/science/article/pii/S016926071730408X}.

\bibitem[Sekiya et~al.(2019)Sekiya, Sakaino, and Toshiaki]{8637973}
Masashi Sekiya, Sho Sakaino, and Tsuji Toshiaki.
\newblock Linear logistic regression for estimation of lower limb muscle activations.
\newblock \emph{IEEE Transactions on Neural Systems and Rehabilitation Engineering}, 27\penalty0 (3):\penalty0 523--532, 2019.
\newblock \doi{10.1109/TNSRE.2019.2898207}.

\bibitem[Zaman et~al.(2022)Zaman, Xiang, Rakshit, and Yang]{9546647}
Rahid Zaman, Yujiang Xiang, Ritwik Rakshit, and James Yang.
\newblock Hybrid predictive model for lifting by integrating skeletal motion prediction with an opensim musculoskeletal model.
\newblock \emph{IEEE Transactions on Biomedical Engineering}, 69\penalty0 (3):\penalty0 1111--1122, 2022.
\newblock \doi{10.1109/TBME.2021.3114374}.

\bibitem[McErlain-Naylor et~al.(2021)McErlain-Naylor, King, and Felton]{app11041450}
Stuart~A. McErlain-Naylor, Mark~A. King, and Paul~J. Felton.
\newblock A review of forward-dynamics simulation models for predicting optimal technique in maximal effort sporting movements.
\newblock \emph{Applied Sciences}, 11\penalty0 (4), 2021.
\newblock ISSN 2076-3417.
\newblock \doi{10.3390/app11041450}.
\newblock URL \url{https://www.mdpi.com/2076-3417/11/4/1450}.

\bibitem[Kim et~al.(2009)Kim, Fernandez, Akbarshahi, Walter, Fregly, and Pandy]{https://doi.org/10.1002/jor.20876}
Hyung~J. Kim, Justin~W. Fernandez, Massoud Akbarshahi, Jonathan~P. Walter, Benjamin~J. Fregly, and Marcus~G. Pandy.
\newblock Evaluation of predicted knee-joint muscle forces during gait using an instrumented knee implant.
\newblock \emph{Journal of Orthopaedic Research}, 27\penalty0 (10):\penalty0 1326--1331, 2009.
\newblock \doi{https://doi.org/10.1002/jor.20876}.
\newblock URL \url{https://onlinelibrary.wiley.com/doi/abs/10.1002/jor.20876}.

\bibitem[Modenese et~al.(2011)Modenese, Phillips, and Bull]{MODENESE20112185}
L.~Modenese, A.T.M. Phillips, and A.M.J. Bull.
\newblock An open source lower limb model: Hip joint validation.
\newblock \emph{Journal of Biomechanics}, 44\penalty0 (12):\penalty0 2185--2193, 2011.
\newblock ISSN 0021-9290.
\newblock \doi{https://doi.org/10.1016/j.jbiomech.2011.06.019}.
\newblock URL \url{https://www.sciencedirect.com/science/article/pii/S0021929011004647}.

\bibitem[Rane et~al.(2019)Rane, Ding, McGregor, and Bull]{RN96}
Lance Rane, Ziyun Ding, Alison~H. McGregor, and Anthony M.~J. Bull.
\newblock Deep learning for musculoskeletal force prediction.
\newblock \emph{Annals of Biomedical Engineering}, 47\penalty0 (3):\penalty0 778--789, 2019.
\newblock ISSN 1573-9686.
\newblock \doi{10.1007/s10439-018-02190-0}.
\newblock URL \url{https://doi.org/10.1007/s10439-018-02190-0}.

\bibitem[Trinler et~al.(2019)Trinler, Schwameder, Baker, and Alexander]{TRINLER201955}
Ursula Trinler, Hermann Schwameder, Richard Baker, and Nathalie Alexander.
\newblock Muscle force estimation in clinical gait analysis using anybody and opensim.
\newblock \emph{Journal of Biomechanics}, 86:\penalty0 55--63, 2019.
\newblock ISSN 0021-9290.
\newblock \doi{https://doi.org/10.1016/j.jbiomech.2019.01.045}.
\newblock URL \url{https://www.sciencedirect.com/science/article/pii/S002192901930082X}.

\bibitem[Lee et~al.(2009)Lee, Narayanan, Kannan, Mendel, and Krovi]{4912337}
Leng-Feng Lee, Madusudanan~S. Narayanan, Srikanth Kannan, Frank Mendel, and Venkat~N. Krovi.
\newblock Case studies of musculoskeletal-simulation-based rehabilitation program evaluation.
\newblock \emph{IEEE Transactions on Robotics}, 25\penalty0 (3):\penalty0 634--638, 2009.
\newblock \doi{10.1109/TRO.2009.2019780}.

\bibitem[Lund et~al.(2015)Lund, Andersen, de~Zee, and Rasmussen]{doi:10.1080/23335432.2014.993706}
Morten~Enemark Lund, Michael~Skipper Andersen, Mark de~Zee, and John Rasmussen.
\newblock Scaling of musculoskeletal models from static and dynamic trials.
\newblock \emph{International Biomechanics}, 2\penalty0 (1):\penalty0 1--11, 2015.
\newblock \doi{10.1080/23335432.2014.993706}.
\newblock URL \url{https://doi.org/10.1080/23335432.2014.993706}.

\bibitem[Campanini et~al.(2020)Campanini, Disselhorst-Klug, Rymer, and Merletti]{10.3389/fneur.2020.00934}
Isabella Campanini, Catherine Disselhorst-Klug, William~Z. Rymer, and Roberto Merletti.
\newblock Surface emg in clinical assessment and neurorehabilitation: Barriers limiting its use.
\newblock \emph{Frontiers in Neurology}, 11, 2020.
\newblock ISSN 1664-2295.
\newblock \doi{10.3389/fneur.2020.00934}.
\newblock URL \url{https://www.frontiersin.org/articles/10.3389/fneur.2020.00934}.

\bibitem[Teramae et~al.(2018)Teramae, Noda, and Morimoto]{8004462}
Tatsuya Teramae, Tomoyuki Noda, and Jun Morimoto.
\newblock Emg-based model predictive control for physical human–robot interaction: Application for assist-as-needed control.
\newblock \emph{IEEE Robotics and Automation Letters}, 3\penalty0 (1):\penalty0 210--217, 2018.
\newblock \doi{10.1109/LRA.2017.2737478}.

\bibitem[Zhao et~al.(2020{\natexlab{a}})Zhao, Zhang, Li, Yang, Dehghani-Sanij, and Xie]{9258965}
Yihui Zhao, Zhiqiang Zhang, Zhenhong Li, Zhixin Yang, Abbas~A. Dehghani-Sanij, and Shengquan Xie.
\newblock An emg-driven musculoskeletal model for estimating continuous wrist motion.
\newblock \emph{IEEE Transactions on Neural Systems and Rehabilitation Engineering}, 28\penalty0 (12):\penalty0 3113--3120, 2020{\natexlab{a}}.
\newblock \doi{10.1109/TNSRE.2020.3038051}.

\bibitem[Buchanan et~al.(2004)Buchanan, Lloyd, Manal, and Besier]{Buchanan2004NeuromusculoskeletalME}
Thomas~S. Buchanan, David~G. Lloyd, Kurt Manal, and Thor~F. Besier.
\newblock Neuromusculoskeletal modeling: estimation of muscle forces and joint moments and movements from measurements of neural command.
\newblock \emph{Journal of applied biomechanics}, 20 4:\penalty0 367--95, 2004.

\bibitem[Simonetti et~al.(2021)Simonetti, Koopman, and Sartori]{RN84}
D.~Simonetti, Bfjm Koopman, and M.~Sartori.
\newblock Clusterization of multi-channel electromyograms into muscle-specific activations to drive a subject-specific musculoskeletal model: towards fast and accurate clinical decision-making.
\newblock \emph{Annu Int Conf IEEE Eng Med Biol Soc}, 2021:\penalty0 5979--5982, 2021.
\newblock ISSN 2375-7477.
\newblock \doi{10.1109/embc46164.2021.9631016}.
\newblock 2694-0604 Simonetti, D Koopman, B F J M Sartori, M Journal Article Research Support, Non-U.S. Gov't United States 2021/12/12 Annu Int Conf IEEE Eng Med Biol Soc. 2021 Nov;2021:5979-5982. doi: 10.1109/EMBC46164.2021.9631016.

\bibitem[Silvestros et~al.(2019)Silvestros, Preatoni, Gill, Gheduzzi, Hernandez, Holsgrove, and Cazzola]{10.1371/journal.pone.0216663}
Pavlos Silvestros, Ezio Preatoni, Harinderjit~S. Gill, Sabina Gheduzzi, Bruno~Agostinho Hernandez, Timothy~P. Holsgrove, and Dario Cazzola.
\newblock Musculoskeletal modelling of the human cervical spine for the investigation of injury mechanisms during axial impacts.
\newblock \emph{PLOS ONE}, 14\penalty0 (5):\penalty0 1--20, 05 2019.
\newblock \doi{10.1371/journal.pone.0216663}.
\newblock URL \url{https://doi.org/10.1371/journal.pone.0216663}.

\bibitem[Su et~al.(2021)Su, Qi, Li, Chen, Ferrigno, and De~Momi]{9380441}
Hang Su, Wen Qi, Zhijun Li, Ziyang Chen, Giancarlo Ferrigno, and Elena De~Momi.
\newblock Deep neural network approach in emg-based force estimation for human–robot interaction.
\newblock \emph{IEEE Transactions on Artificial Intelligence}, 2\penalty0 (5):\penalty0 404--412, 2021.
\newblock \doi{10.1109/TAI.2021.3066565}.

\bibitem[Wu et~al.(2021)Wu, Cao, Fei, Yang, Xu, Zhang, Zeng, and Song]{10.1371/journal.pone.0247883}
Changcheng Wu, Qingqing Cao, Fei Fei, Dehua Yang, Baoguo Xu, Guanglie Zhang, Hong Zeng, and Aiguo Song.
\newblock Optimal strategy of semg feature and measurement position for grasp force estimation.
\newblock \emph{PLOS ONE}, 16\penalty0 (3):\penalty0 1--21, 03 2021.
\newblock \doi{10.1371/journal.pone.0247883}.
\newblock URL \url{https://doi.org/10.1371/journal.pone.0247883}.

\bibitem[Burton et~al.(2021)Burton, Myers, and Rullkoetter]{BURTON2021110439}
William~S. Burton, Casey~A. Myers, and Paul~J. Rullkoetter.
\newblock Machine learning for rapid estimation of lower extremity muscle and joint loading during activities of daily living.
\newblock \emph{Journal of Biomechanics}, 123:\penalty0 110439, 2021.
\newblock ISSN 0021-9290.
\newblock \doi{https://doi.org/10.1016/j.jbiomech.2021.110439}.
\newblock URL \url{https://www.sciencedirect.com/science/article/pii/S0021929021002190}.

\bibitem[Zhang et~al.(2023)Zhang, Zhao, Shone, Li, Frangi, Xie, and Zhang]{9970372}
Jie Zhang, Yihui Zhao, Fergus Shone, Zhenhong Li, Alejandro~F. Frangi, Sheng~Quan Xie, and Zhi-Qiang Zhang.
\newblock Physics-informed deep learning for musculoskeletal modeling: Predicting muscle forces and joint kinematics from surface emg.
\newblock \emph{IEEE Transactions on Neural Systems and Rehabilitation Engineering}, 31:\penalty0 484--493, 2023.
\newblock \doi{10.1109/TNSRE.2022.3226860}.

\bibitem[Shi et~al.(2023)Shi, Ma, Zhao, and Zhang]{shi2023physicsinformed}
Yue Shi, Shuhao Ma, Yihui Zhao, and Zhiqiang Zhang.
\newblock A physics-informed low-shot learning for semg-based estimation of muscle force and joint kinematics, 2023.

\bibitem[Pau et~al.(2012)Pau, Xie, and Pullan]{6226835}
James W.~L. Pau, Shane S.~Q. Xie, and Andrew~J. Pullan.
\newblock Neuromuscular interfacing: Establishing an emg-driven model for the human elbow joint.
\newblock \emph{IEEE Transactions on Biomedical Engineering}, 59\penalty0 (9):\penalty0 2586--2593, 2012.
\newblock \doi{10.1109/TBME.2012.2206389}.

\bibitem[Buchanan et~al.()Buchanan, Lloyd Dg Fau~Manal, Manal K Fau~Besier, and Besier]{RN78}
T.~S. Buchanan, Kurt Lloyd Dg Fau~Manal, Thor~F. Manal K Fau~Besier, and T.~F. Besier.
\newblock Neuromusculoskeletal modeling: estimation of muscle forces and joint moments and movements from measurements of neural command.
\newblock \penalty0 (1065-8483 (Print)).

\bibitem[Millard et~al.(2013)Millard, Uchida, Seth, and Delp]{10.1115/1.4023390}
Matthew Millard, Thomas Uchida, Ajay Seth, and Scott~L. Delp.
\newblock {Flexing Computational Muscle: Modeling and Simulation of Musculotendon Dynamics}.
\newblock \emph{Journal of Biomechanical Engineering}, 135\penalty0 (2):\penalty0 021005, 02 2013.
\newblock ISSN 0148-0731.
\newblock \doi{10.1115/1.4023390}.
\newblock URL \url{https://doi.org/10.1115/1.4023390}.

\bibitem[Lloyd and Besier(2003)]{LLOYD2003765}
David~G Lloyd and Thor~F Besier.
\newblock An emg-driven musculoskeletal model to estimate muscle forces and knee joint moments in vivo.
\newblock \emph{Journal of Biomechanics}, 36\penalty0 (6):\penalty0 765--776, 2003.
\newblock ISSN 0021-9290.
\newblock \doi{https://doi.org/10.1016/S0021-9290(03)00010-1}.
\newblock URL \url{https://www.sciencedirect.com/science/article/pii/S0021929003000101}.

\bibitem[Thelen(2003)]{10.1115/1.1531112}
Darryl~G. Thelen.
\newblock {Adjustment of Muscle Mechanics Model Parameters to Simulate Dynamic Contractions in Older Adults }.
\newblock \emph{Journal of Biomechanical Engineering}, 125\penalty0 (1):\penalty0 70--77, 02 2003.
\newblock ISSN 0148-0731.
\newblock \doi{10.1115/1.1531112}.
\newblock URL \url{https://doi.org/10.1115/1.1531112}.

\bibitem[Zhao et~al.(2020{\natexlab{b}})Zhao, Zhang, Li, Yang, Dehghani-Sanij, and Xie]{article}
Yihui Zhao, Zhiqiang Zhang, Zhenhong Li, Zhixin Yang, AA~Dehghani-Sanij, and Sheng Xie.
\newblock An emg-driven musculoskeletal model for estimating continuous wrist motion.
\newblock \emph{IEEE Transactions on Neural Systems and Rehabilitation Engineering}, PP, 11 2020{\natexlab{b}}.
\newblock \doi{10.1109/TNSRE.2020.3038051}.

\bibitem[Zajac(1989)]{Zajac1989MuscleAT}
Felix~E. Zajac.
\newblock Muscle and tendon: properties, models, scaling, and application to biomechanics and motor control.
\newblock \emph{Critical reviews in biomedical engineering}, 17 4:\penalty0 359--411, 1989.
\newblock URL \url{https://api.semanticscholar.org/CorpusID:25888803}.

\bibitem[Ramsay et~al.(2009)Ramsay, Hunter, and Gonzalez]{RAMSAY2009463}
John~W. Ramsay, Betsy~V. Hunter, and Roger~V. Gonzalez.
\newblock Muscle moment arm and normalized moment contributions as reference data for musculoskeletal elbow and wrist joint models.
\newblock \emph{Journal of Biomechanics}, 42\penalty0 (4):\penalty0 463--473, 2009.
\newblock ISSN 0021-9290.
\newblock \doi{https://doi.org/10.1016/j.jbiomech.2008.11.035}.
\newblock URL \url{https://www.sciencedirect.com/science/article/pii/S0021929008006003}.

\bibitem[Zhang et~al.(2020)Zhang, Guo, and Zanotto]{8930581}
Huanghe Zhang, Yi~Guo, and Damiano Zanotto.
\newblock Accurate ambulatory gait analysis in walking and running using machine learning models.
\newblock \emph{IEEE Transactions on Neural Systems and Rehabilitation Engineering}, 28\penalty0 (1):\penalty0 191--202, 2020.
\newblock \doi{10.1109/TNSRE.2019.2958679}.

\bibitem[Saul et~al.(2015)Saul, Hu, Goehler, Vidt, Daly, Velisar, and Murray]{RN86}
K.~R. Saul, X.~Hu, C.~M. Goehler, M.~E. Vidt, M.~Daly, A.~Velisar, and W.~M. Murray.
\newblock Benchmarking of dynamic simulation predictions in two software platforms using an upper limb musculoskeletal model.
\newblock \emph{Comput Methods Biomech Biomed Engin}, 18\penalty0 (13):\penalty0 1445--58, 2015.
\newblock ISSN 1025-5842 (Print) 1025-5842.
\newblock \doi{10.1080/10255842.2014.916698}.
\newblock 1476-8259 Saul, Katherine R Hu, Xiao Goehler, Craig M Vidt, Meghan E Daly, Melissa Velisar, Anca Murray, Wendy M R01 EB011615/EB/NIBIB NIH HHS/United States R24 HD050821/HD/NICHD NIH HHS/United States Journal Article England 2014/07/06 Comput Methods Biomech Biomed Engin. 2015;18(13):1445-58. doi: 10.1080/10255842.2014.916698. Epub 2014 Jul 4.

\end{thebibliography}






\end{document}